\documentclass[nouppercase]{kaia}
\usepackage{amsmath}
\title{Generating High-Resolution Regional Precipitation Using Conditional Diffusion Model}

\author[2]{Naufal Shidqi}{a}
\author[2]{Chaeyoon Jeong}{a}
\author[2]{Sungwon Park}{a}
\author[4]{Elke Zeller}{c}
\author[4]{Arjun Babu Nellikkattil}{c}
\author{Karandeep Singh}{b}
\affiliation{School of Computing, KAIST, Daejeon, Republic of Korea, \{naufal.shidqi, lily9991, psw0416\}@kaist.ac.kr}{a}
\affiliation{Data Science Group, IBS, Daejeon, Republic of Korea, \{ksingh\}@ibs.re.kr}{b}
\affiliation{Department of Climate System, PNU, Busan, Republic of Korea, \{arjunbabun, elkezeller\}@pusan.ac.kr}{c}
\affiliation{Center for Climate Physics, IBS, Busan, Republic of Korea}{d}

\usepackage{graphicx} 
\usepackage{amsmath}
\usepackage{amsfonts}
\usepackage{algorithm}
\usepackage{algpseudocode}
\usepackage{xcolor}

\begin{document}

\maketitle
\begin{abstract}
	Climate downscaling is a crucial technique within climate research, serving to project low-resolution (LR) climate data to higher resolutions (HR). Previous research has demonstrated the effectiveness of deep learning for downscaling tasks. However, most deep learning models for climate downscaling may not perform optimally for high scaling factors (i.e., 4x, 8x) due to their limited ability to capture the intricate details required for generating HR climate data. Furthermore, climate data behaves differently from image data, necessitating a nuanced approach when employing deep generative models. In response to these challenges, this paper presents a deep generative model for downscaling climate data, specifically precipitation on a regional scale. We employ a denoising diffusion probabilistic model (DDPM) conditioned on multiple LR climate variables. The proposed model is evaluated using precipitation data from the Community Earth System Model (CESM) v1.2.2 simulation. Our results demonstrate significant improvements over existing baselines, underscoring the effectiveness of the conditional diffusion model in downscaling climate data.
\end{abstract}

\section{Introduction}

\begin{figure}[t]
\begin{center}
\includegraphics[width=1.0\linewidth]{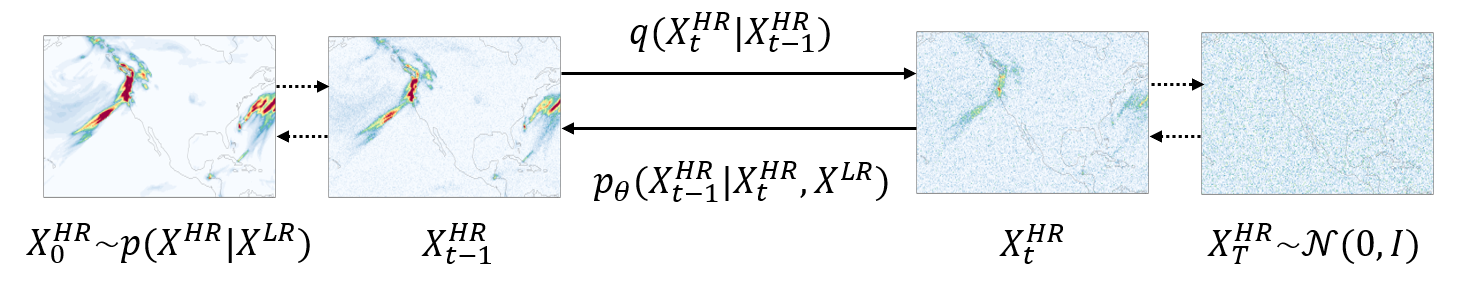}
\end{center}
   \caption{Overview of forward diffusion process by adding Gaussian noise to a single climate variable (total precipitation, PRECT) of HR data ($q$, left to right) and reverse diffusion process by gradually denoising the noisy data ($p_{\theta}$, right to left) conditioned on $X^{LR}$.}

\label{fig:diffprocess}
\end{figure}

The Earth's climate is undergoing rapid changes, and its impact on the planet is becoming increasingly evident. The buildup of greenhouse gases in the atmosphere, mainly from human activities like burning fossil fuels, and deforestation, is causing climate change. It has significant and far-reaching effects on Earth. For instance, increased heatwave frequency and intensity, extreme weather, and altered precipitation patterns brought about by rising global temperatures harm agriculture and ecosystems. Therefore, it is crucial to get a better understanding of the climate system. Climate researchers achieve this by employing both global and local climate models, which utilize fundamental equations to simulate the climate system at both regional and global scales.
Despite the significance of high-resolution simulations in benefiting climate prediction and analysis \cite{iles2020benefits}, there are practical constraints related to computational requirements and the associated costs. While global climate models are readily available online through various agencies, local climate models take more resources. ``Downscaling'' (a term that refers to projecting global climate data, down onto regional scales) is a cost-effective method of approximating high-resolution data.

Climate modelers often use one of two existing approaches: dynamical downscaling and statistical downscaling. Dynamical downscaling involves utilizing high-resolution (HR) regional climate models to simulate the intricate interactions between various elements, including the atmosphere, land, oceans, and other climate variables. In essence, this is running a climate model on a regional scale with a high resolution. This method offers a more comprehensive representation of local climate conditions but necessitates substantial computational resources \cite{small2014new}. On the other hand, statistical downscaling leverages climate data to identify statistical relationships between HR climate models and observed local climate conditions. Conventional statistical techniques often struggle to capture the intricate relationships between global and local climate patterns and the interactions among climate variables. Statistical downscaling can be viewed as a form of super-resolution (SR) task, analogous to those encountered in computer vision research. Similar to how SR tasks enhance the resolution of low-resolution images, statistical downscaling enhances the resolution of coarse climate data.

Prior research has established the promise of deep learning in the context of downscaling climate models. One work has employed a deep SR convolutional neural network to facilitate the downscaling of climate variables \cite{vandal2017deepsd}. Other investigations have explored alternative deep learning techniques, including LSTM and various deep neural network-based methods, aiming to generate localized climate data by incorporating climate-physics characteristics into their models \cite{chou2021generating, park2022downscaling}. However, it's worth noting that these models predominantly rely on regression approaches \cite{saharia2022image}, which may exhibit limitations in handling higher scale factors, i.e., 4x and 8x,  and could result in high-frequency information loss due to over-smoothing, a consequence of their objective functions. On the contrary, the GAN-based method \cite{cheng2020generating} has emerged as an alternative solution for climate data downscaling. Nevertheless, it's essential to acknowledge that the training of GANs can be complex, requiring efforts to achieve convergence and the inclusion of additional discriminator models. Furthermore, the success of super-resolution deep learning models in computer vision does not always seamlessly transfer to the domain of climate data and warrants careful consideration.

To address the limitations of previous approaches mentioned above, this paper presents a denoising diffusion probabilistic model (DDPM) for downscaling climate data. The model is conditioned on multiple low-resolution (LR) climate variables to extract the intrinsic relationships between the input variables (LR) and the target variable (HR). Moreover, we experiment with how the number of climate variables used in the model affects the performance, highlighting the need for a careful approach when employing deep learning in climate downscaling. We evaluate our model by downscaling total precipitation (PRECT) from the Community Earth System Model (CESM) v1.2.2 into high 4x and 8x scale factors. The results show substantial improvements in the performance compared to existing baselines, indicating the effectiveness of the conditional diffusion model in downscaling climate data. These findings pave the way for future research endeavors involving the use of diffusion models for the downscaling of climate data from a wide range of climate model simulations.

\begin{figure}[t]
\begin{center}
\includegraphics[width=1.0\linewidth]{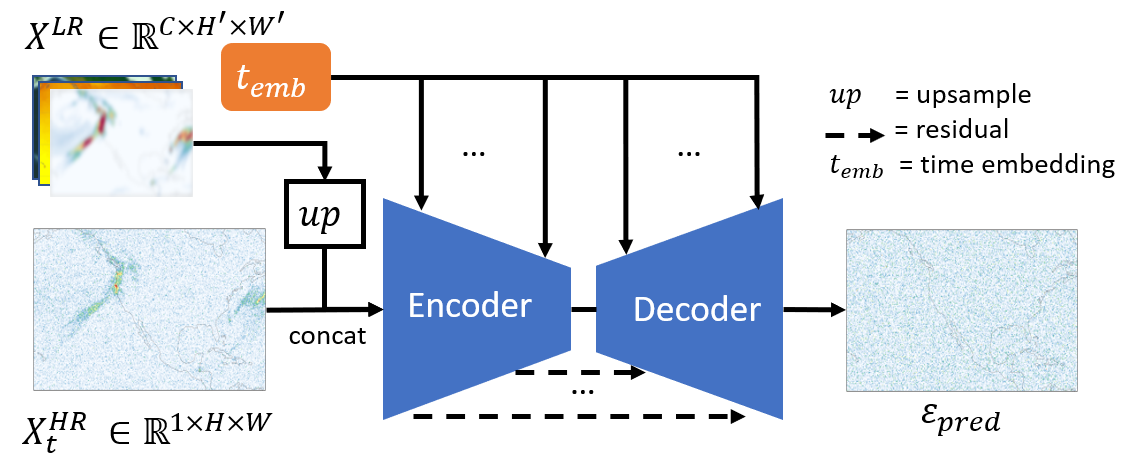}
\end{center}
   \caption{Overall architecture of the denoising model $\epsilon_{\theta}$ based on U-Net \cite{ronneberger2015u} used in DDPM \cite{ho2020denoising}, conditioned on three LR climate variables $X^{LR}$ (TS, PRECT, and dPHIS). The model predicts only the noise injected ($\epsilon_{pred}$ ) on a single target climate variable (PRECT) at timestep $t$.}
\label{fig:ddpm_unet}
\end{figure}

\section{Conditional Diffusion Model}

Our objective is to produce high-resolution (HR) regional climate data from low-resolution (LR) global climate data, a process referred to as “climate downscaling”. More specifically, given LR global-scale climate data represented as $X^{LR} \in \mathbb{R}^{C \times H^{'} \times W^{'}}$ and HR regional-scale climate data denoted as $X^{HR} \in \mathbb{R}^{C \times H \times W}$ originating from an unknown conditional distribution $p(X^{HR}|X^{LR})$, we adapt denoising diffusion probabilistic model (DDPM) \cite{ho2020denoising} for generating the downscaled climate data $X^{HR}$ in an iterative improvement procedure. The model initiates the generation of the downscaled output $X_{0}^{HR}$ through a reverse diffusion process in $T$ steps, starting from a noisy data  $X_{T}^{HR} \sim \mathcal{N}(0, \mathbf{I})$, to intermediate data distribution $X_{t}^{HR}$ and progressing iteratively until reaching $X_{0}^{HR}$ where $t$ takes on values from $T-1$ to 1. The learnable reverse diffusion process  is conditioned on $X^{LR}$, denoted as $p(X_{t-1}^{HR}|X_{t}^{HR}, X^{LR})$ where $X_{0}^{HR} \sim p(X^{HR}|X^{LR})$, as illustrated in Figure \ref{fig:diffprocess}. The intermediate data distributions at each diffusion step $X_{t}^{HR}$ are obtained from a forward diffusion process which is determined beforehand by adding fixed Gaussian noise iteratively using Markov chain, denoted as $q(X_{t}^{HR}|X_{t-1}^{HR})$  which can be expressed as:

\begin{equation}
q(X_{t}^{HR}|X_{t-1}^{HR}) = \mathcal{N}(X_{t}^{HR}|\sqrt{1-\beta_{t}}X_{t-1}^{HR}, \beta_{t}\textbf{I})	
\end{equation}

\noindent where $\beta_{t}$ is a pre-defined noise variance scheduling hyperparameter. On the other hand, the learnable reverse Gaussian diffusion process, which is also Markov chain, can be mathematically defined as:

\begin{equation}
p_{\theta}(X_{t-1}^{HR}|X_{t}^{HR}, X^{LR}) = \mathcal{N}(X_{t-1}^{HR}|\mu_{\theta}(X^{LR}, X_{t}^{HR}, t), \sigma_{t}^{2}\textbf{I})	
\end{equation}

\subsection{Denoising Model Training and Inference}
We follow DDPM \cite{ho2020denoising} framework as our denoising model with U-Net \cite{ronneberger2015u} architecture as the neural network backbone as shown in Figure \ref{fig:ddpm_unet}. The overall algorithm for training the model and model inference is described in Algorithm \ref{alg:training} and Algorithm 2 respectively. 

\begin{algorithm}
\caption{Model Training}
\label{alg:training}
\begin{algorithmic}[1]
\Repeat
    \State sample $(X^{LR}, X_{0}^{HR}) \sim p(X^{LR}, X^{HR})$
    \State sample $t \sim Uniform({1, . . ., T})$
    \State sample $\epsilon \sim \mathcal{N}(0, \mathbf{I})$
    \State upsample $X^{LR}$ to match $X^{HR}$ size
    \State take gradient descent step on
        \State \hspace{1em} $\nabla_\theta=\parallel\epsilon-\epsilon_{\theta}(X^{LR}, X_{t}^{HR}, t)\parallel$,
        \State \hspace{1em }with $X_t^{HR}=\sqrt{\Bar{\alpha_t}}X_0^{HR}+\sqrt{1-\Bar{\alpha_t}}\epsilon$ 

\Until converged
\end{algorithmic}
\end{algorithm}

The denoising model objective is to estimate the initial noise $\epsilon \sim \mathcal{N}(0, \mathbf{I})$ by minimalizing the objective function L1 loss, $\mathcal{L}$:

\begin{equation}
    \mathcal{L}=\mathbb{E}_{X^{LR}, X_{t}^{HR}, \epsilon, t}\parallel\epsilon-\epsilon_{\theta}(X^{LR}, \sqrt{\Bar{\alpha_t}}X_0^{HR}+\sqrt{1-\Bar{\alpha_t}}\epsilon, t)\parallel
\end{equation}

\noindent where $\epsilon_\theta$ is the denoising model, $\alpha_t=1-\beta_t$, and $\Bar{\alpha_t}=\prod_{s=1}^t \alpha_s$. Then the trained $\epsilon_\theta$ is used to iteratively denoise a pure noisy data $X_{T}^{HR} \sim \mathcal{N}(0, \mathbf{I})$ to generate the output $X_0^{HR}$, as summarized in Algorithm \ref{alg:infer}.

\begin{algorithm}
\caption{Model Inference}
\label{alg:infer}
\begin{algorithmic}[1]

\State sample $X_T^{HR} \sim \mathcal{N}(0, \mathbf{I})$
\State upsample $X^{LR}$ to match $X^{HR}$ size
\For{$t = T, ...,1$}
    \State sample $z \sim \mathcal{N}(0, \mathbf{I})$ \textbf{if} {$t>1$} \textbf{else}{$z=0$}
    \State $X_{t-1}^{HR} = \frac{1}{\sqrt{\alpha_t}}(X_t^{HR} - \frac{1-\alpha_t}{\sqrt{1-\Bar{\alpha_t}}} \epsilon_{\theta}(X^{LR}, X_{t}^{HR}, t)\ + \sigma_t z)$
\EndFor
\State \Return $X_0^{HR}$
\end{algorithmic}
\end{algorithm}

\subsection{Configuring Climate Variables}
To configure the climate variables in the denoising model, whether as conditioning input variables ($X^{LR}$) or as target variables ($X^{HR}$), one can implement it by simply using all of the variables as both conditioning input variables and target variables, i.e., $X^{LR} \in \mathbb{R}^{C \times H^{'} \times W^{'}}$ and  $X^{HR} \in \mathbb{R}^{C \times H \times W}$. However, in later experiments, we discovered that using all of the variables as conditioning input and target output is not as effective as focusing on only one variable as the model output, i.e.,  $X^{LR} \in \mathbb{R}^{C \times H^{'} \times W^{'}}$ and  $X^{HR} \in \mathbb{R}^{1 \times H \times W}$ instead. Following the approach in \cite{ho2020denoising, saharia2022image}, the conditioning variables $X^{LR}$ are concatenated with the $X_{t}^{HR}$ along the channel dimension, as illustrated in Figure \ref{fig:ddpm_unet}. 
\section{Experiment Setup}
\subsection{Data}
For our experiments, we use Surface Temperature (TS), Total Precipitation (PRECT), and the gradient of Topography (dPHIS) climate variables from a global climate HR simulation conducted with the fully-coupled, global circulation climate model  CESM version 1.2.2. The HR data from the climate simulation data ($X^{HR} \in \mathbb{R}^{C \times H \times W}$) represents daily means over the North America region for 20 years ($N = 20\times365 = 7300$). We split the data into 5300 datapoints for the training, 500 for the validation, and 1500 for the test data. The HR data has a resolution of 213$\times$321, with each pixel representing an approximate grid resolution of 25 kilometers. We perform center cropping to 192$\times$256 resolution and normalization to zero means. To generate LR data ($X^{LR} \in \mathbb{R}^{C \times H^{'} \times W^{'}}$), we apply bicubic interpolation with antialiasing enabled, as in \cite{saharia2022image}, to convert the HR data into a coarser resolution ($H^{'} \times W^{'}$), i.e. into 4x and 8x scale factors. Subsequently, we upsample this LR data to match the target output size ($H \times W$) for the model’s input. We use TS, PRECT, and dPHIS from the LR data as the conditioning input variables of the model and set PRECT as the target variable for the model output. 

\subsection{Training and Evaluation Details}


We train the model in a supervised setting with $X^{LR}$ and $X^{HR}$ pairs, as described in the previous section. Then we perform model inference on $X^{LR}$ from test dataset as a condition to the model to generate downscaled output of $X_0^{HR}$.

\noindent \textbf{Model Configuration.} The model employs uniform convolution kernels with a size of 3 throughout the network and maintains five levels of feature map resolution in both the encoder (reduced by half at each level) and decoder (doubled at each level) components of the U-Net. The depth of the feature map channel is set to 128 in the first and second levels, while it is doubled in the third and fifth levels. During the diffusion process, we configure the number of timesteps $T$ to be 100 and apply a simple linear $\beta$ schedule, as detailed in \cite{ho2020denoising}. 

\noindent \textbf{Training Details.} Model training is conducted using the Adam optimizer \cite{KingmaB14} and a cosine annealing learning rate scheduler with an initial learning rate of 2E-5. All models, including baselines, undergo 265,000 training iterations, with the final checkpoint chosen for evaluation. The U-Net baseline employs the same network architecture, and we mostly adhere to the configuration outlined in \cite{park2022downscaling, ledig2017photo} for the SRResNet baseline. 

\noindent \textbf{Evaluation Details.} In our evaluation, we downscale 1500 data points extracted from the CESM v1.2.2 climate dataset. For our model, we employ three climate variables as LR conditioning input to downscale precipitation as target output (3in1out). Furthermore, we also conduct experiments that downscale all three climate variables (3in3out) to assess the impact of the number of projecting variables employed in the model. To ensure a fair comparison, we apply the same variable input and output configurations (3in3out, 3in1out) to both the SRResNet and U-Net baselines. Performance assessment is carried out by calculating the root mean squared error (RMSE) through a comparison between the model's output and the high-resolution ground truth data over 4x and 8x scale factors.
\section{Result and Discussion}

\begin{table}
\begin{center}
\begin{tabular}{c|c|cc}
\hline
\multirow{2}{*}{\textbf{Model}} & \textbf{In-Out} & \multicolumn{2}{c}{\textbf{PRECT}} \\ \cline{3-4} 
 & \textbf{Config} & \textbf{4x} & \textbf{8x} \\ 
 \hline \hline
Bilinear & \textbf{-} & 4.0389 & 5.2660 \\ \hline
Bicubic & \textbf{-} & 4.0235 & 5.3193 \\ \hline
SRResNet & 3in3out & 4.1063 & 5.2005 \\ \hline
SRResNet & 3in1out & 4.1536 & 5.1980 \\ \hline
U-Net & 3in3out & 4.2775 & 5.7003 \\ \hline
U-Net & 3in1out & 4.2121 & 5.4501 \\ \hline
Cond. DDPM & 3in3out & 4.1438 & 5.5015 \\ \hline
Cond. DDPM (Ours) & 3in1out & \textbf{3.3447} & \textbf{5.1803} \\ \hline
\end{tabular}
\end{center}
\caption{The RMSE (lower is better) comparison of conventional methods and deep learning models for 4x and 8x (units are 1e-8). The In-Out Config column indicates the number of climate variables used as the model's input and the downscaled output.}
\label{table:rmse}
\end{table}

\begin{figure*}[ht]
\begin{center}
\includegraphics[width=1.0\linewidth]{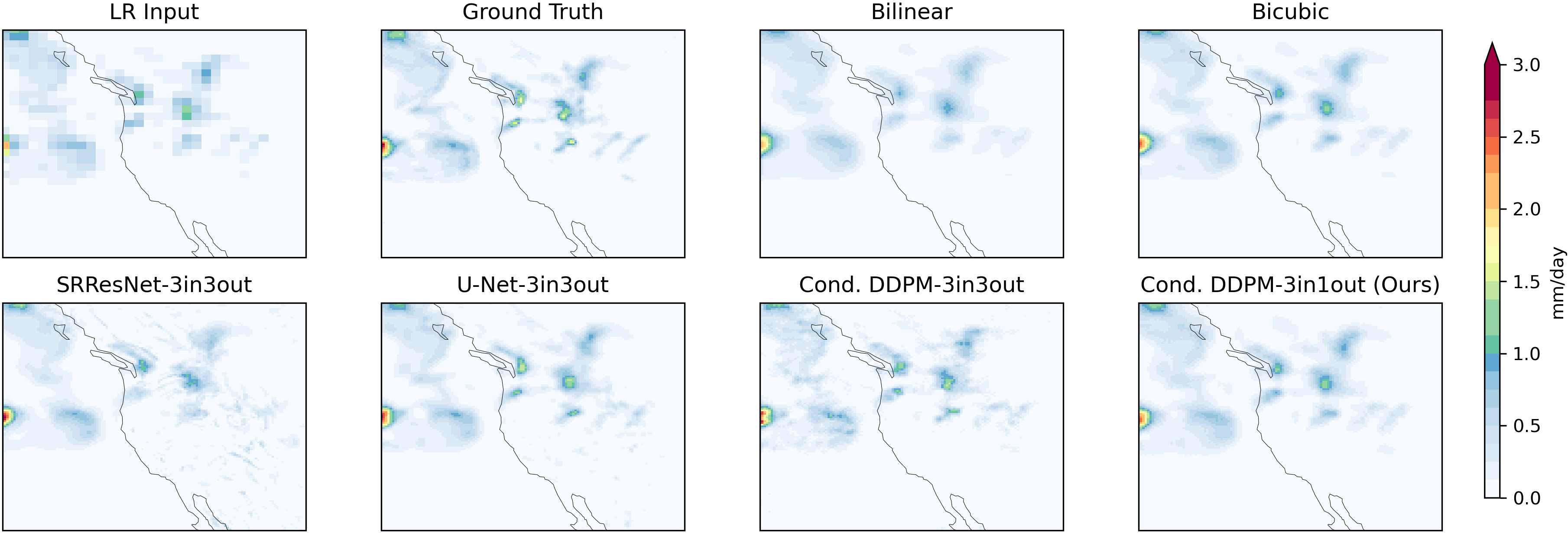}
\end{center}
   \caption{Comparison of downscaling results in a regional scale (4x scale factor) of total precipitation (PRECT) from LR input, ground truth, bilinear, bicubic, SRResNet, U-Net, Conditional DDPM (Cond. DDPM) with 3in3out variables and 3in1out variables configuration, tested on the same datapoint.}
\label{fig:output4x}
\end{figure*}

\noindent \textbf{Quantitative Results.} Table~\ref{table:rmse} presents the results of the model's performance in downscaling PRECT data for high-scale factors of 4x and 8x. Our evaluation includes a comparison with traditional techniques, specifically bilinear and bicubic interpolation, alongside state-of-the-art convolution network models like SRResNet \cite{ledig2017photo} and U-Net \cite{ronneberger2015u}. Bilinear and bicubic interpolation are well-established, commonly employed methods for downscaling climate data, while SRResNet and U-Net represent deep convolution networks commonly employed in computer vision tasks.

It is evident that our model with 3in1out configuration surpasses the baseline performance for both scale factors. In the case of the 4x scale factor, our model enhances performance by reducing RMSE by 16.87\% compared to bicubic interpolation and by 20.59\% compared to the U-Net with the same configuration as its underlying network. Furthermore, when dealing with the higher 8x scale factors, our model still demonstrates superior performance in comparison to the other models, although the improvements are not as significant as in the 4x scale factor. This could be influenced by the model's architecture configuration, where we maintain consistency in the number of residual layers, levels of feature map resolution, and channel depth multipliers for both scale factors. Nonetheless, the superior performance of our model underscores the effectiveness of employing the denoising diffusion model for downscaling, particularly when generating one variable at a time and utilizing multiple climate data as conditioning variables.

\noindent \textbf{Qualitative Results.} We further delved into the qualitative results by visually examining the outputs of downscaling at the 4x scale factor, as depicted in Figure~\ref{fig:output4x}. Each image provides a closer view of the North America region, with colors indicating varying levels of total precipitation. Notably, bilinear and bicubic methods yield outputs that are considerably blurry, albeit preserving the essence of the LR input. In contrast, when comparing the deep learning baselines with our model, SRResNet and U-Net introduce undesired artifacts in their outputs. This occurrence may be attributed to the handling of inter-channel relationships, which mirrors practices within the field of computer vision (using 3in3out configuration). A similar pattern can be observed in conditional DDPM (Cond. DDPM) when using the 3in3out configuration, where some distinct details emerge when compared to the ground truth. In contrast, our model (3in1out) produces an accurate representation of the LR input, with only minor deviations from the ground truth. This observation underscores the need for a more careful approach to the design and training of inter-channel and intra-channel relationships in downscaling climate variables. This requirement is empirically addressed by employing the 3in1out configurations.

\section{Conclusion and Future Work}

In this research, we introduce a deep generative model designed to efficiently downscale a global-scale climate variable to a regional scale. Our model employs a denoising diffusion model conditioned on low-resolution (LR) climate variables, yielding a single target climate variable, as opposed to generating an equivalent number of variables as the model input (similar to image channels in computer vision applications). We substantiate the model's effectiveness through extensive evaluations involving the total precipitation (PRECT) variable extracted from the Community Earth System Model (CESM) v1.2.2. This method presents a promising avenue for downscaling climate data and underscores the potential of generative diffusion models in this context. 

The current work still has limitations that future endeavors may address by exploring customized denoising model architectures tailored for climate science. It's important to recognize that climate variables cannot be treated in the same way as channels in the image domain. Incorporating recent advances in DDPM \cite{song2020denoising, nichol2021improved} will be also a promising subsequent work to improve the model inference process. Of course, this should still involve careful consideration from the perspective of climate variables. Additionally, there's an opportunity to utilize the diffusion model to generate high-resolution climate data from disparate global (LR) and regional (HR) climate data sources, which will be a primary focus of our forthcoming research. This study also serves as a catalyst for further investigations in climate research, including the examination of historical climate data, the prediction of future climate patterns, and the mitigation of anomalies in climate events.
\section*{Acknowledgement}
The climate simulations were conducted on the IBS/ICCP supercomputer “Aleph”, a 1.43 petaflops high-performance Cray XC50-LC Skylake computing system.  This work was supported by the Institute for Basic Sciences (IBS), Republic of Korea, under IBS-R029-C2 and IBS-R028-D1.

\bibliography{references}

\newpage
\newpage


\end{document}